\documentclass[sn-apa]{sn-jnl}

\usepackage{graphicx}%
\usepackage{multirow}%
\usepackage{amsmath,amssymb,amsfonts}%
\usepackage{amsthm}%
\usepackage[title]{appendix}%
\usepackage{xcolor}%
\usepackage{textcomp}%
\usepackage{manyfoot}%
\usepackage{booktabs}%
\usepackage{algorithm}%
\usepackage{algorithmicx}%
\usepackage{algpseudocode}%
\usepackage{listings}%

\usepackage{longtable}
\usepackage{makecell}
\usepackage{multirow}
\usepackage{multicol}
\usepackage{siunitx}

\begin{document}

\title[Article Title]{Bridging Gaps in Hate Speech Detection: Meta-Collections and Benchmarks for Low-Resource Iberian Languages}


\author[1]{\fnm{Paloma} \sur{Piot}}\email{paloma.piot@udc.es}

\author[2]{\fnm{José Ramom} \sur{Pichel Campos}}\email{jramon.pichel@usc.gal}

\author[1]{\fnm{Javier} \sur{Parapar}}\email{javier.parapar@udc.es}

\affil[1]{\orgdiv{IRLab, CITIC Research Centre}, \orgname{Universidade da Coruña}, \orgaddress{\country{Spain}}}

\affil[2]{\orgdiv{CiTIUS}, \orgname{Universidade de Santiago de Compostela}, \orgaddress{\country{Spain}}}


\abstract{
Hate speech poses a serious threat to social cohesion and individual well-being, particularly on social media, where it spreads rapidly. While research on hate speech detection has progressed, it remains largely focused on English, resulting in limited resources and benchmarks for low-resource languages. Moreover, many of these languages have multiple linguistic varieties, a factor often overlooked in current approaches. At the same time, large language models require substantial amounts of data to perform reliably, a requirement that low-resource languages often cannot meet. In this work, we address these gaps by compiling a meta-collection of hate speech datasets for European Spanish, standardised with unified labels and metadata. This collection is based on a systematic analysis and integration of existing resources, aiming to bridge the data gap and support more consistent and scalable hate speech detection. We extended this collection by translating it into European Portuguese and into a Galician standard that is more convergent with Spanish and another Galician variant that is more convergent with Portuguese, creating aligned multilingual corpora. Using these resources, we establish new benchmarks for hate speech detection in Iberian languages. We evaluate state-of-the-art large language models in zero-shot, few-shot, and fine-tuning settings, providing baseline results for future research. Moreover, we perform a cross-lingual analysis with our target languages. Our findings underscore the importance of multilingual and variety-aware approaches in hate speech detection and offer a foundation for improved benchmarking in underrepresented European languages.
}




\maketitle

\section{Introduction}\label{intro}

Hate speech remains a significant issue on social media, posing serious threats to individual safety and community cohesion \citep{GonzlezBailn2022}. The rapid expansion of online platforms has facilitated the spread of harmful content, amplifying its impact and making moderation increasingly challenging \citep{pewresearch2021onlineharassment, hickey2023auditing}. Recent research shows that approximately 30\% of young individuals face cyberbullying \citep{kansok2023systematic}, while 46\% of Black/African American adults have reported online racial harassment \citep{ADL2024}, highlighting the critical importance of tackling hate speech. At the same time, some platforms have scaled back human content moderation efforts, often prioritising user engagement over strict enforcement of community guidelines \citep{theguardianMetaFactcheckers}, which has further intensified the spread of harmful speech. Consequently, this reduction in human oversight creates an urgent need for effective detection systems capable of identifying and mitigating hate speech promptly across all languages and linguistic varieties.

However, most research on hate speech detection has focused on high-resource languages, particularly English, where large-scale datasets and advanced language models have driven significant progress \citep{fortuna2018survey,Vidgen2020,PiotMartinRodillaParapar2024}. In contrast, low-resource languages remain largely underrepresented, with limited annotated data to train effective models. To overcome this, current approaches often depend on cross-lingual transfer learning and multilingual Large Language Models (LLMs), which leverage knowledge from high-resource languages to support low-resource ones \citep{ranasinghe-zampieri-2023-text}. Although multilingual LLMs have demonstrated potential for handling multiple languages \citep{zhong2024opportunitieschallengeslargelanguage}, their performance in low-resource contexts is still inconsistent, especially for languages with substantial regional variation \citep{alam-etal-2024-llms,joshi-etal-2025-adapting}. This gap underscores the need to evaluate state-of-the-art LLMs in underrepresented linguistic settings and to develop tailored resources that enhance hate speech detection for such languages.

Beyond the lack of annotated datasets and linguistic resources, low-resource languages present additional challenges due to internal linguistic diversity. Spanish and Portuguese, for example, have distinct European and American varieties. And as for Galician, a Romance language spoken in north-western Spain (Galicia and other communities), it is extremely close to Portuguese, if it uses the same orthography, or to Castilian, with its current orthography \citep{Pichel2020}, which can only converge with Spanish or Portuguese, as Carvalho Calero points out \citep{calero1981problemas}. While some datasets explicitly label texts by variety \citep{castillo-lopez-etal-2023-analyzing}, this distinction is often not taken into account during model training, with data from different varieties being merged or used interchangeably. This can lead to suboptimal performance when models are applied to specific regional varieties.

To better understand the current landscape, we examined available hate speech datasets for the languages and varieties of interest. For Portuguese, available datasets are limited and focus on Brazilian Portuguese, leaving European Portuguese largely unrepresented \citep{leite-etal-2020-toxic,vargas-etal-2022-hatebr,Trajano2023,salles-etal-2025-hatebrxplain}. The situation is somewhat better for Spanish, where several publicly available datasets exist. These include a mix of European and Latin American Spanish, with some explicitly distinguishing between the two. In contrast, Galician lacks any dedicated hate speech datasets. The only related resource is a misogyny detection corpus derived from one of the Spanish datasets included in our review \citep{alvarez-crespo-castro-2024-galician}. This gap highlights the need for targeted datasets that better represent European Portuguese, European Spanish, and Galician.

\subsection{Research questions}

To guide our work, we formulate the following research questions:

\begin{enumerate}
    \item \textit{\textbf{RQ1}: To what extent do existing hate speech datasets reflect the linguistic characteristics of European Spanish, European Portuguese, and Galician?}
    \item \textit{\textbf{RQ2}: Can synthetic data generation effectively compensate for the lack of annotated hate speech resources in European Portuguese and Galician?}
    \item \textit{\textbf{RQ3}: How well do state-of-the-art large language models perform in detecting hate speech across these low-resource linguistic varieties?}
    \item \textit{\textbf{RQ4}: Does accounting for Galician's internal variation (Spanish-like vs. Portuguese-like) affect model performance?}
\end{enumerate}

To address these questions, this study investigates the representation of European Spanish, European Portuguese, and Galician in current hate speech resources and explores strategies to improve coverage and performance in these settings. Guided by our research questions, we analyse existing datasets, generate synthetic data for low-resourced languages, and evaluate the performance of state-of-the-art large language models. Our findings underscore the pressing need for dedicated hate speech resources in European Portuguese and Galician. While translated and synthetic datasets, especially those derived from Spanish, offer meaningful improvements, tailored language-specific data remains critical for achieving optimal results. Fine-tuning consistently outperforms zero and few-shot approaches, reinforcing the importance of targeted dataset development. By centring linguistic diversity and low-resource challenges, we provide new benchmarks and insights to advance hate speech detection in the Iberian context.

\subsection{Contributions}

In this work, we address these challenges by focusing on hate speech detection in three underrepresented linguistic varieties: European Spanish, European Portuguese, and Galician. Our contributions are five-fold. First, we conduct a systematic review of existing hate speech datasets in Spanish, Portuguese, and Galician, selecting those that align with our definition of hate speech and the target language varieties. Based on this review, we construct a meta-collection of relevant resources. Second, to address the lack of data for European Portuguese and Galician, we generate synthetic datasets to support further research in these languages. Third, we perform a detailed lexical and linguistic analysis of the datasets, highlighting both their strengths and current limitations within the Iberian context. Fourth, we evaluate state-of-the-art large language models under zero-shot, few-shot, and fine-tuning settings, establishing new benchmarks for hate speech detection in Iberian languages. Moreover, we evaluate the cross-lingual performance of our target languages. Finally, we introduce a novel perspective by addressing the linguistic duality of Galician, distinguishing between its Spanish and Portuguese-influenced variants, a factor not previously explored in machine learning or natural language processing for hate speech detection.

\section{Related work}\label{related-work}

\subsection{Hate speech detection}
Detecting hate speech is a challenging task that involves not only technical complexity but also ethical and social considerations. The boundaries between offensive language and hate speech are often blurred, shaped by context, cultural norms, and individual perception \citep{davidsonhatespeech,pamungkas-etal-2020-really}. Issues such as subjectivity, freedom of expression, and evolving language (e.g., slang or coded speech) further complicate consistent annotation and detection \citep{waseem-hovy-2016-hateful}. This phenomena has been extensively studied across a range of computational frameworks, evolving from traditional machine learning approaches to more recent deep learning architectures. Early studies relied on methods such as Logistic Regression and Support Vector Machines \citep{waseem-hovy-2016-hateful, Davidson_Warmsley_Macy_Weber_2017, Chatzakou2017, Tahmasbi2018}, often leveraging handcrafted linguistic features.

Subsequent advances in representation learning introduced deep neural networks, including Convolutional Neural Networks (CNNs) and Recurrent Neural Networks (RNNs), which improved performance by automatically extracting semantic patterns from text \citep{qian-etal-2019-benchmark}. After that, transformer-based language models such as BERT \citep{devlin-etal-2019-bert, grimminger-klinger-2021-hate} and RoBERTa \citep{glavas-etal-2020-xhate} have become the dominant paradigm, showing promising results across multiple hate speech benchmarks. Recent work has shown that LLMs now represent the state of the art in hate speech detection, achieving strong results even in few-shot and zero-shot settings. These models demonstrate the ability to generalise across domains and linguistic contexts with minimal supervision, offering a promising alternative to traditional supervised approaches \citep{plaza-del-arco-etal-2023-respectful, roy-etal-2023-probing, wang2022toxicitydetectiongenerativepromptbased}.

In parallel, there is growing interest in multimodal approaches that integrate textual, visual, and user-level metadata to better capture the multifaceted nature of online hate \citep{Perifanos2021, Yang2022}. These methods are particularly relevant in real-world applications, where hate speech is often embedded in complex social and multimedia contexts.

\subsection{Datasets} 

The availability of annotated datasets is crucial for the development and evaluation of hate speech detection systems. While early efforts were predominantly centered on English \citep{Poletto2020,Vidgen2020,PiotMartinRodillaParapar2024}, recent studies have expanded their focus to include a broader range of languages and dialects. Notable contributions include datasets for Spanish \citep{Fersini2018, basile-etal-2019-semeval, s19214654}, Portuguese \citep{fortuna-etal-2019-hierarchically}, Italian \citep{sanguinetti-etal-2018-italian, evalita2020}, French and Arabic \citep{ousidhoum-etal-2019-multilingual}, Turkish \citep{toraman-etal-2022-large}, and Slovene \citep{10.1007/978-3-030-27947-9_9}.

Despite this progress, most available datasets still concentrate on high-resource or globally dominant languages. As a result, European Portuguese, one of the 24 official languages of the European Union,  and Galician, a co-official language in an autonomous region (Galicia), continue to be significantly under-represented \citep{craith2006european}. This limits the applicability of current models in multilingual and cross-cultural settings and underscores the need for targeted resource development in these contexts.

\subsection{Synthetic data generation methods}

Synthetic data generation has become an essential approach to address the lack of annotated resources, especially for low-resource and underrepresented languages. Key strategies include machine translation from high-resource languages, controlled text generation using language models, and various data augmentation techniques.

Machine translation is a common method for generating parallel datasets, enabling the projection of annotations from a well-resourced source language into the target language \citep{Pamungkas2021}. While this method supports large-scale data creation, it may introduce stylistic biases or translation artifacts, especially when source and target languages differ significantly in structure or cultural context.

Complementary approaches use rule-based or generative techniques to create hate speech content from scratch. The CONAN dataset \citep{chung-etal-2019-conan} employs crowd-sourcing combined with adversarial examples to simulate realistic, context-aware hate speech. Meanwhile, the Dynamically Generated Hate Speech dataset \citep{vidgen-etal-2021-learning} uses controlled text generation to produce a wide variety of toxic language expressions by conditioning outputs on hate categories and targets of the hate. These datasets enable systematic exploration of hate speech across nuanced dimensions and support more robust evaluation of detection systems.

More recently, instruction-tuned LLMs have been leveraged to produce synthetic hate speech and non-hate content. These models can be prompted with predefined templates, scenarios, or seed phrases to create synthetic training instances that reflect specific linguistic or contextual features \citep{jahan2024comprehensivestudynlpdata,Girn2025}. While effective in simulating realistic content, this approach raises ethical and methodological challenges, including the risk of reinforcing biases and the need for careful validation of generated outputs.

\section{Systematic dataset collection}\label{method}

This section describes the methodology used to identify and collect hate speech datasets for European Spanish, European Portuguese, and Galician. The process included four steps: (1) defining keywords, (2) searching digital libraries and repositories, (3) filtering resources using set criteria, and (4) extracting relevant datasets. The dataset collection was conducted in December 2024, and included papers published up to November 2024. The following subsections describe each step in detail.

\subsection{Keywords}\label{sec:method_keywords}

To identify relevant datasets, we selected keywords commonly associated with hate speech detection. The term ``hate speech'' is related to concepts such as cyberbullying, abusive language, offensive language, and specific types of hate speech (e.g., racism, sexism, misogyny). Based on this, we used the following keywords: \texttt{hate speech}, \texttt{hateful}, \texttt{abusive}, \texttt{offensive}, \texttt{toxic}, \texttt{aggressive}, \texttt{cyberbullying}, \texttt{sexism}, \texttt{misogyny}, and \texttt{racism}.  

To narrow the search to our target languages, we combined these terms with ``Spanish'', ``Portuguese'', and ``Galician'', along with ``dataset'' or ``corpus''. We did not specify language variants (e.g., European vs. Latin American Spanish) at this stage to maximise recall. This filtering was done later.

\subsection{Sources} \label{sec:method_sources}

To identify relevant datasets, we searched across multiple digital libraries and dataset repositories. This covered both academic publications and curated repositories focused on hate speech datasets, ensuring broad coverage of available resources.

For academic publications, we queried the following digital libraries:

\begin{itemize}
    \item \textbf{ACL Anthology}: a widely used resource in natural language processing research, particularly for computational linguistics and machine learning applications.
    \item \textbf{ACM Digital Library}: a repository containing peer-reviewed computer science research, including works on dataset creation.
    \item \textbf{Scopus}: a large abstract and citation database that indexes publications from various disciplines, including computer science and social sciences.
\end{itemize}

We used the keyword combinations described in Section \ref{sec:method_keywords} to query each library. Table \ref{tab:queries} lists the specific queries and the number of retrieved documents per source. To refine the results, we applied additional filters: in Scopus, we limited the subject areas to ``Computer Science'', ``Social Sciences'', and ``Engineering'', and included only document types classified as conference ``papers'', ``articles'', ``book chapters'', or ``data papers''. In the ACM Digital Library, we restricted results to ``research articles''.

In addition to academic sources, we examined specialised repositories focused on hate speech datasets. These repositories included:
\begin{itemize}
    \item \textbf{Hate Speech Data}\footnote{\url{https://hatespeechdata.com}}: a curated repository of publicly available datasets for hate speech detection \citep{Vidgen2020}.
    \item \textbf{Hate Speech Supersets}: an aggregated resource that compiling multilingual hate speech datasets \citep{tonneau-etal-2024-languages}.
\end{itemize}

Unlike academic digital libraries, these repositories are dedicated to datasets and do not rely on keyword-based searches. Instead, we manually reviewed their catalogues to identify datasets in our target languages. Table \ref{tab:repositories} shows the number of datasets meeting our criteria in each repository.

\begin{table}[htb]
    \centering
    \caption{Queries run against the digital libraries and documents retrieved.}
    \begin{tabular}{p{1.5cm} p{9cm} r}
        \toprule
        \textbf{Source} & \textbf{Query} & \textbf{Docs.} \\ \midrule
        \textbf{ACL Anthology} & \texttt{(title:hate* OR title:abuse* OR title:offens* OR title:toxic* OR title:aggress* OR title:cyberbully* OR title:sexism* OR title:misogyny* OR title:racism* OR abstract:hate* OR abstract:abuse* OR abstract:offens* OR abstract:toxic* OR abstract:aggress* OR abstract:cyberbully* OR abstract:sexism* OR abstract:misogyny* OR abstract:racism*) AND (title:dataset* OR title:corpus* OR abstract:dataset* OR abstract:corpus*) AND (Spanish OR Portuguese OR Galician)} & 666 \\ 
        \textbf{ACM Digital Library} & \texttt{[[All: abstract.] AND [[All: "hate speech*"] OR [All: "abuse*"] OR [All: "offens*"] OR [All: "toxic*"] OR [All: "aggress*"] OR [All: "cyberbully*"] OR [All: "sexism*"] OR [All: "misogyny*"] OR [All: "racism*"]] AND [All: abstract.] AND [[All: "dataset*"] OR [All: "corpus*"]] AND [All: abstract.] AND [[All: "spanish"] OR [All: "portuguese"] OR [All: "galician"]]] OR [[All: title.] AND [[All: "hate speech*"] OR [All: "abuse*"] OR [All: "offens*"] OR [All: "toxic*"] OR [All: "aggress*"] OR [All: "cyberbully*"] OR [All: "sexism*"] OR [All: "misogyny*"] OR [All: "racism*"]] AND [All: title.] AND [[All: "dataset*"] OR [All: "corpus*"]] AND [All: title.] AND [[All: "spanish"] OR [All: "portuguese"] OR [All: "galician"]]]} & 308 \\
        \textbf{Scopus} & \texttt{TITLE-ABS-KEY( "hate speech*" OR "hateful*" OR "abuse*" OR "offens*" OR "toxic*" OR "aggress*" OR "cyberbully*" OR "sexism*" OR "misogyny*" OR "racism*" ) AND TITLE-ABS-KEY( "dataset*" OR "corpus*" ) AND TITLE-ABS-KEY( "Spanish" OR "Portuguese" OR "Galician")} & 272 \\
        \bottomrule
    \end{tabular}
    \label{tab:queries}
\end{table}

\begin{table}[htb]
    \centering
    \caption{Matching results in repositories for our target languages.}
    \begin{tabular}{lr}
        \toprule
        \textbf{Source} &  \textbf{Docs.} \\ \midrule
        \textbf{Hate Speech Data} &  7 \\ 
        \textbf{Hate Speech Supersets} &  9 \\
        \bottomrule
    \end{tabular}
    \label{tab:repositories}
\end{table}

\subsection{Criteria} \label{sec:method_criteria}

After retrieving documents from the digital libraries and repositories, we manually assessed each for inclusion in our dataset collection based on the following criteria:

\begin{enumerate}
    \item The dataset must be \textbf{new}, introducing a resource not previously published.
    \item It must contain \textbf{text from social networks}, excluding images, audio, or content from non-social media sources such as books or song lyrics.
    \item It must be \textbf{annotated by humans} to ensure labelling is reliable.
    \item It's \textbf{definition of hate speech} must align with ours or cover related categories such as racism or sexism. We define hate speech as ``\textit{language characterised by offensive, derogatory, humiliating, or insulting discourse \citep{fountahatespeech} that promotes violence, discrimination, or hostility towards individuals or groups \citep{davidsonhatespeech} based on attributes such as race, religion, ethnicity, or gender \citep{hatelingo2018elsherief,ElSherief2018,hatemm2023}}'', which aligns closely with the United Nations' definition \citep{unhatespeech}.
    \item It must target \textbf{European Spanish, European Portuguese, or Galician}, to ensure linguistic relevance.
\end{enumerate}

Documents not meeting these criteria were excluded. This process ensured the relevance and consistency of the dataset collection with our research objectives.

The filtering process was conducted in stages. First, we excluded papers that did not introduce a new dataset or that included languages outside our target set. We also removed datasets not derived from social media platforms. After this initial filtering, 28 Spanish datasets, 13 Portuguese datasets, and 1 Galician dataset remained.

Next, we performed a detailed review of each dataset, filtering out those that did not meet the following criteria: (1) human annotation, (2) alignment with our hate speech definition, and (3) use of the European variants of Spanish and Portuguese. The final number of datasets per language is presented in Table \ref{tab:count_datasets}.

Following dataset identification, we attempted to access the data. As shown in Table \ref{tab:count_datasets}, we successfully obtained 10 datasets for European Spanish, either through publicly availability, email requests or authorisation. However, no datasets for European Portuguese or Galician were accessible. Some datasets lacked access links, and attempts to contact the authors were unsuccessful. Consequently, for this study, we focus on the datasets available at the time of analysis.

These findings address \textit{\textbf{RQ1}: To what extent do existing hate speech datasets reflect the linguistic characteristics of European Spanish, European Portuguese, and Galician?} While multiple datasets were identified and accessed for European Spanish, none were accessible for European Portuguese or Galician. This indicates that among the three languages, only European Spanish is represented to a meaningful extent, though, still considerably less than more widely resourced languages such as English.

\begin{table}[htb]
    \centering
    \caption{Matching results in repositories for our target languages.}
    \begin{tabular}{lrr}
        \toprule
        \textbf{Language} &  \textbf{Fitting Datasets} & \textbf{Available Datasets} \\ \midrule
        \textbf{European Spanish} &  21 & 10 \\ 
        \textbf{European Portuguese} &  3 & 0 \\
        \textbf{Galician} &  1 & 0 \\
        \bottomrule
    \end{tabular}
    \label{tab:count_datasets}
\end{table}

\section{Corpus creation}\label{corpus}

This section describes the construction of the meta-collection corpus (MetaHateES) and the subsequent generation of synthetic data for European Portuguese and Galician based on this source.

\subsection{Meta-collection construction}

As mentioned, our systematic dataset collection resulted in ten available European Spanish datasets. These datasets, listed below, form the basis of the new meta-collection, MetaHateES.

\subsubsection{DETESTS 2024 \citep{detests2024}} The DETESTS dataset was constructed to detect and classify racial stereotypes in Spanish, incorporating annotations that capture both explicit and implicit stereotypes. It consists of two text sources: comments on news articles (from DETESTS and NewsCom-TOX) and tweets reacting to racial hoaxes (from StereoHoax-ES). The dataset is based on online platforms, including Spanish news outlets (ABC, elDiario.es, El Mundo, NIUS, Men\'eame) and Twitter (now X), focusing on discussions related to immigration. DETESTS primarily targets racial stereotypes, often linked to xenophobia and misinformation. The dataset comprises \num{12111} texts (\num{6762} from DETESTS and \num{5349} from StereoHoax-ES). Authors define stereotype as \textit{``cognitive mechanism consisting of a set of beliefs regarding another social group, namely the outgroup, which is perceived as different. Forming these beliefs involves a homogenisation process, where one characteristic of an individual or part of the group is generalised to the entire group. This generalisation typically occurs based on factors such as place of origin, ethnicity, or religion. Stereotypes can be expressed explicitly or implicitly''}.

\subsubsection{EXIST 2021 \& 2023 \citep{exist2022}} The EXIST dataset was constructed to detect and categorise sexism in social media, focusing on Twitter and Gab. It includes posts labelled for binary sexism identification and multi-class categorisation into direct, judgmentally, objectifying, ideological, or violent sexism. Data from 2021 includes around \num{5700} Twitter and Gab posts, while the 2023 dataset consists of \num{4209} Twitter posts with annotations for sexism presence (Task 1: ``YES'') and type (Task 2: ``DIRECT'' or ``JUDGMENTAL''). The 2024 dataset is identical to 2023. Authors based their work on sexism defined as in the Oxford English Dictionary as \textit{``prejudice, stereotyping or discrimination, typically against women, on the basis of sex''}.

\subsubsection{HaSCoSVa 2023 \citep{castillo-lopez-etal-2023-analyzing}} The HaSCoSVa dataset is a hate speech dataset comprising different variants of Spanish. Their targets are immigrants and the data is from Twitter, consisting of 4000 instances. It contains labels in a binary fashion and it also contains its variation (latam or europe), where 2500 are from Spanish European. In their paper, the report the differences between Latin American Spanish and European Spanish, and how different models identify it differently. Authors do not provide a formal definition of what constitutes hate speech, but based on their work and data, their study includes misogyny and xenophobia, two hate speech constructs, that they define as \textit{``harm against women due to gender, which might result in psychological, reputational, professional, or even physical damage''} and \textit{``attitudes, prejudices, and behaviour that reject, exclude and often vilify persons, based on the perception that they are outsiders or foreigners to the community, society or national identity''}, respectively.

\subsubsection{SemEval HatEval 2019 \citep{basile-etal-2019-semeval}} The SemEval HatEval dataset is about detecting hate speech against immigrants and women in both Spanish and English. It is part of SemEval 2019 Task 5, and they define one task for binary hate speech detection an another for identifying other features in hateful content. The dataset contains is from Twitter and contains 19600 tweets, of which 6600 are in Spanish. They use Castilian Spanish annotators, in order to have native speakers for the content. Authors follow the hate speech definition \textit{``any communication that disparages a person or a group on the basis of some characteristic such as race, color, ethnicity, gender, sexual orientation, nationality, religion, or other characteristics''}.

\subsubsection{Hate Football 2023 \citep{hatefootball2023}} This corpus is from comments from Twitter around the topic of Spanish men's football league. The dataset contains 7483 tweets and is labelled on aggressive, misogyny, racist, and safe. We considered data points labelled as misogyny and racist as hate speech, as their definition of aggressiveness does not fall under the umbrella of our definition of hate speech, but misogynist and racist does. Authors follow Cambridge Dictionary's definition of hate speech: \textit{``public speech that expresses hate or encourages violence towards a person or group based on something such as race, religion, sex, or sexual orientation''}.

\subsubsection{HaterNet 2019 \citep{PereiraKohatsu2019}} HaterNet consists on a dataset of 6000 instances for hate speech detection in Spanish. It is labelled in a binary fashion. The dataset includes a diverse range of hate-related content, reflecting real-world discourse on social media. Authors define hate speech as \textit{``a kind of speech that denigrates a person or multiple persons based on their membership to a group, usually defined by race, ethnicity, sexual orientation, gender identity, disability, religion, political affiliation, or views''}.

\subsubsection{MeTwo 2020 \citep{9281090}} The MeTwo dataset consists of sexist expressions and attitudes, including subtle forms, in Twitter in Spanish. It is focused on misogynistic and xenophobic content, and was curated using an specialised lexicon. The expressions used for the creation are in European Spanish. Authors use \cite{davidsonhatespeech} definition of hate speech \textit{``language that is used to expresses hatred towards a targeted group or is intended to be derogatory, to humiliate, or to insult the members of the group''}.

\subsubsection{MisoCorpus 2021 \citep{GARCIADIAZ2021506}} The MisoCorpus is a dataset for detecting misogynous messages on Twitter. It is classified in violence towards women, messages harassing women and general traits related to misogyny. The dataset is focused on both Latin American Spanish and European Spanish. We filtered the data to keep only the post published in Spain, leaving us with 1527 data instances. Authors use \cite{manne2017down} definition of misogyny: \textit{``social environments in which women will tend to face hostility of various kinds because they are women in a man's world who are held to be failing to live up to men's standards''}.

\subsubsection{NewsCom-TOX 2024 \citep{Taul2024}} The NewsCom-TOX dataset contains 4359 comments in Spanish posted on responses to news articles related to immigration. Their objective was to identify messages with racial and linguistic categories, specially toxicity and its intensity. Although toxicity may be considered different as hate speech, the authors define it as \textit{``comment that attacks, hurts, threatens, insults, offends, denigrates or disqualifies a person or group of people on the basis of characteristics such as race, ethnicity, nationality, political ideology, religion, gender and sexual orientation, among others, regardless of whether the writer intends to be hurtful or offensive''}, which matches our definition of hate speech. 

\subsubsection{OffendES 2021 \citep{plaza-del-arco-etal-2021-offendes}} The OffendES corpus comprises 47128 Spanish-language comments sourced from Twitter, Instagram, and YouTube, focusing on interactions involving young influencers. Each comment is manually annotated across multiple predefined offensive categories, with a subset including confidence scores for each label. This design supports both multi-class classification and multi-output regression analyses. Authors follow the offensive definition of \textit{``the text which uses hurtful, derogatory, or obscene terms made by one person to another person''}. They claim that related terms include hate speech, toxic language, aggressive language and abusive language and that, although there are subtle differences in meaning, they are all compatible with the above general definition.

\subsection{Standardise data} \label{sec:method_standard}

This section describes the procedures used to standardise the datasets for the meta-collection. To unify the diverse annotations, we applied mappings and transformations to consolidate all labels into a single consistent format.

Each dataset was loaded individually, retaining only columns relevant to the meta-collection. Rows with missing values and duplicates were removed to ensure data quality.

To align labelling, we applied mapping strategies: labels such as ``No'', ``False'', and ``NO'' (non-hate content) were standardised to 0; labels like ``True'', ``Offensive'', ``Sexist'', ``Aggressive'', and shorthand forms such as ``OFG'' or ``OFP'' were mapped to 1, indicating hate speech. When a gold label was absent, functions inferred the final label by majority voting across annotations.

A new column, \texttt{content\_type}, categorises content types from the original datasets, including \textit{offensive}, \textit{hate speech}, \textit{sexism}, \textit{stereotype}, \textit{aggressive}, \textit{toxicity}, and \textit{misogyny}. These categories represent subconstructs within our hate speech definition. Each dataset was assigned a category based on its definitions or label naming. This classification provides insight into the types of hate speech represented and improves the granularity of future analysis. Additionally, each entry was tagged with its \texttt{source}, indicating the data origin, such as \textit{Twitter}, \textit{YouTube}, \textit{news outlets}, \textit{Instagram}, and \textit{Gab}.

\subsection{Dataset statistics}

The meta-collection contains \num{82233} samples, with 24.85\% labelled as hate speech and 75.15\% as non-hate. The data come from five sources and ten distinct datasets.

Table~\ref{tab:sources} presents the number of hate speech samples, hate rate (\%), and total samples for each source. Twitter is the largest source, contributing \num{43254} samples and with a hate rate of 29.70\%. YouTube contributes \num{22538} samples with a hate rate of 16.94\%. News sources account for \num{9983} samples at 27.50\%. Instagram and Gab contribute \num{5968} and \num{490} samples, with hate rates of 12.72\% and 54.08\%, respectively.

\begin{table}[ht]
    \small
    \centering
    \renewcommand{\arraystretch}{1.1}
    \setlength{\tabcolsep}{7pt}
    \caption{Distribution of hate speech samples, hate rates, and total samples across different platforms.}
    \begin{tabular}{@{}lrrrr@{}}
        \toprule
        \textbf{Source} & \textbf{Hate Samples} & \textbf{Hate Rate (\%)} & \textbf{Total Samples} \\
        \midrule
        \texttt{Gab}        & \num{265} & 54.08\%   & \num{490} \\
        \texttt{Twitter}    & \num{12848} & 29.70\% & \num{43254} \\
        \texttt{News}       & \num{2745} & 27.50\%  & \num{9983} \\
        \texttt{YouTube}    & \num{3819} & 16.94\%  & \num{22538} \\
        \texttt{Instagram}  & \num{759} & 12.72\%   & \num{5968} \\
        \bottomrule
    \end{tabular}
    \label{tab:sources}
\end{table}

Table~\ref{tab:datasets} lists the number of samples per dataset. \texttt{OffendES 2021} is the largest, with \num{30416} samples, followed by \texttt{DETESTS 2024} (\num{9906}) and \texttt{Hate Football 2023} (\num{7482}). The remaining datasets contribute between \num{1527} and \num{6600} samples each.

\begin{table}[ht]
    \small
    \centering
    \renewcommand{\arraystretch}{1.3}
    \setlength{\tabcolsep}{6pt}
    \caption{Number of samples included from each dataset in the meta-collection.}
    \begin{tabular}{@{}l@{\hspace{6pt}}r@{}}
        \toprule
        \textbf{Dataset} & \textbf{Samples} \\
        \midrule
        \texttt{DETESTS 2024}           & \num{9906} \\
        \texttt{EXIST 2021 \& 2023}     & \num{9910} \\
        \texttt{HasCoSVa 2023}          & \num{2500} \\
        \texttt{Hate Football 2023}     & \num{7482} \\
        \texttt{HaterNet 2019}          & \num{5938} \\
        \texttt{MeTwo 2020}             & \num{3600} \\
        \texttt{MisoCorpus 2021}        & \num{1527} \\
        \texttt{NewsCom-TOX 2024}       & \num{4354} \\
        \texttt{OffendES 2021}          & \num{30416} \\
        \texttt{SemEval HateEval 2019}  & \num{6600} \\
        \bottomrule
    \end{tabular}
    \label{tab:datasets}
\end{table}

Table~\ref{tab:content-types} presents the number of hate speech samples, hate rate (\%), and total samples by content type. The ``offensive'' category contains the most samples (\num{30416}), with a hate rate of 15.91\%. ``hate speech'' and ``sexism'' categories have \num{15038} and \num{13510} samples, with hate rates of 30.68\% and 42.31\%, respectively. Other categories include ``stereotype'' (\num{9906} samples, 26.30\%), ``aggressive'' (\num{7482}, 9.26\%), ``toxicity'' (\num{4354}, 31.83\%), and ``misogyny'' (\num{1527}, 38.11\%).

\begin{table}[ht]
    \small
    \centering
    \renewcommand{\arraystretch}{1.3}
    \setlength{\tabcolsep}{8pt}
    \caption{Distribution of hate speech samples and hate rates by annotated content type.}
    \begin{tabular}{@{}lrrr@{}}
        \toprule
        \textbf{Content Type} & \textbf{Hate Samples} & \textbf{Hate Rate (\%)} & \textbf{Total Samples} \\
        \midrule
        \texttt{Offensive}   & \num{4841}  & 15.91\% & 30416 \\
        \texttt{Hate Speech} & \num{4612}  & 30.68\% & 15038 \\
        \texttt{Sexism}      & \num{5717}  & 42.31\% & 13510 \\
        \texttt{Sterotype}   & \num{2606}  & 26.30\% & 9906  \\
        \texttt{Aggresive}   & \num{693}   & 9.26\%  & 7482  \\
        \texttt{Toxicity}    & \num{1386}  & 31.83\% & 4354  \\
        \texttt{Misogyny}    & \num{582}   & 38.11\% & 1527  \\
        \bottomrule
    \end{tabular}
    \label{tab:content-types}
\end{table}

\subsection{New data creation}

This section outlines the process of generating new data in European Portuguese and Galician based on the European Spanish meta-collection (\texttt{MetaHateES}).

\subsubsection{Portuguese}

To create the Portuguese version (\texttt{MetaHatePT}), we focused on producing European Portuguese translations. Since most commercial translation tools default to Brazilian Portuguese, we avoided them due to license access limitation. Instead, we used the \texttt{utter-project/EuroLLM-9B-Instruct} model \citep{martins2024eurollmmultilinguallanguagemodels}, from the EuroLLM project, which supports all European Union languages, including European Portuguese.

We employed the \texttt{easy-translate} framework \citep{garcia-ferrero-etal-2022-model} to interact with the model, with the prompt: ``Translate Spanish to Portuguese (pt\_PT): \%\%SENTENCE\%\%.'' This ensured translations that aligned with European Portuguese linguistic conventions. The resulting dataset, \texttt{MetaHatePT}, retains the same structure and annotation as the original European Spanish version.

\subsubsection{Galician}

To create the Galician variant, we produced two parallel versions from different source languages. The initial version, \texttt{MetaHateGL\_es}, was generated by translating the Spanish meta-collection (MetaHateES) into Galician using Google Translate. Although other online platforms for automatic translation into Galician, such as Opentrad \citep{loinaz2006opentrad} or Proxecto Nós \citep{de2022nos}, are available, we chose Google Translate to maintain consistency for comparisons with future research involving other languages.
This approach enabled efficient translation of large volumes but introduced limitations, such as overly literal translations influenced Spanish grammar and vocabulary patterns. Due to the close linguistic relationship between European Spanish and Galician, translations are often grammatically correct but stylistically marked by European Spanish. For our purposes, this limitation is an advantage, as it produces a Galician variant closely aligned with the European Spanish source, which aligns with our interest in capturing this specific linguistic influence.

The second version, \texttt{MetaHateGL\_pt}, was created by translating the Portuguese dataset (\texttt{MetaHatePT}) into Galician, also using Google Translate. Because the source was Portuguese, these Galician translations show stronger lexical and syntactic similarities to Portuguese. This reflects a more reintegrationist variant of Galician, used in some cultural contexts, differing from the standard Galician common in most digital tools.

By producing both \texttt{MetaHateGL\_es} and \texttt{MetaHateGL\_pt}, we capture internal variation within Galician and enable the analysis of how the translation source (Spanish vs. Portuguese) affects the style and linguistic features of Galician hate speech data.

The data will be made available for research purposes upon request. Since several resources in our meta-collection are subject to individual licensing agreements, researchers will need to secure these permissions to access the full dataset. We will provide guidance to help facilitate the process of obtaining all necessary agreements.

\section{Dataset analysis}\label{analysis}

Now, we present an analysis of the newly developed multilingual resources. We used BERTopic with t-SNE for topic modelling and visualisation. For linguistic analysis, we used spaCy with language-specific models for Spanish and Portuguese. To assess emotional content, we used the NRC Emotion Lexicon, based on Plutchik's emotion model, originally in English and translated to the target languages using Google Translate

\subsection{Lexical analysis}

We started with a basic term frequency analysis in each language. After removing stop words, we extracted the top 20 words. In Spanish, key terms include ``mujer'' (woman), ``puta'' (whore), and ``inmigrante'' (immigrant). Similar patterns appeared in the translated datasets: Portuguese included ``mulher'' (woman), ``puta'' (whore), and ``merda'' (shit); Galician (ES) featured ``muller'' (woman), ``puta'' (whore), and ``inmigrante'' (immigrant); Galician (PT) showed ``muller'' (woman), ``cadela'' (female dog), ``puta'' (whore), and ``país'' (country). The independently sourced English dataset showed different but related terms, including ``fuck'', ``bitch'', ``fagot'', and ``women''. These results indicate a consistent use of gendered and offensive language across languages, especially in datasets derived from the Spanish metacollection.

\begin{figure}[ht]
    \centering
    \includegraphics[scale=0.39]{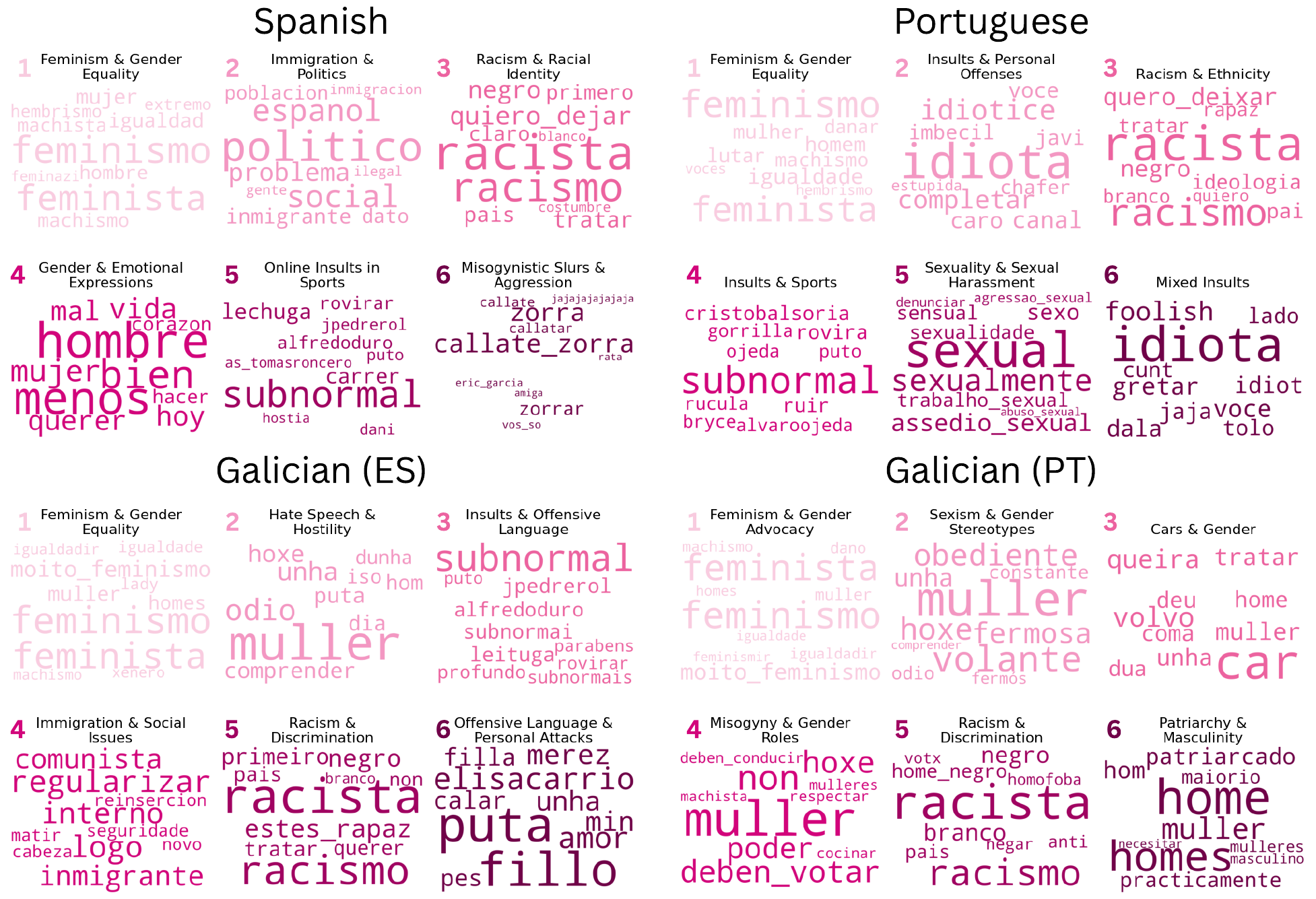}
    \caption{Word clouds of topics for Spanish, Portuguese, Galician (ES) and Galician (PT).}
    \label{fig:wordclouds}
\end{figure}

In our second analysis, we explored our hate speech datasets employing topic modelling techniques. We used the BERTopic framework \citep{grootendorst2022bertopic} with Sentence Transformers to generate sentence embeddings. BERTopic then identified underlying topics in the corpus. We trained the model on each language version to capture key themes in hate speech. Each dataset revealed six major topics. We manually annotated and labelled the topics, enabling a clearer interpretation of the predominant discourses and forms of expression found in the data.  We visualised the top 10 terms in each topic using word clouds (Figure \ref{fig:wordclouds}), and we applied T-distributed Stochastic Neighbour Embedding (t-SNE) \citep{JMLR:v9:vandermaaten08a} to show topic distribution and we visually represent the relationships and proximity of documents within each topic.

For European Spanish, the original metacollection used to create the other language versions, the main topics were: (1) Feminism and Gender Equality, (2) Immigration and Politics, (3) Racism and Racial Identity, (4) Gender and Emotion, (5) Sports Insults, and (6) Misogynistic Slurs. These topics show the range of online speech, including hostile content and social issues.

The Portuguese dataset, translated from Spanish, showed similar categories: (1) Feminism and Gender Equality, (2) Personal Insults, (3) Racism and Ethnicity, (4) Insults and Sports References, (5) Sexual Harassment, and (6) Mixed Insults. Some shifts in topic boundaries were noted, likely due to translation and cultural and linguistic differences between the two languages.

For Galician, translated from both Spanish (ES) and Portuguese (PT), we observed that the Galician (ES) topics include (1) Feminism and Gender, (2) Hate Speech and Hostility, (3) Insults and Offensive Language, (4) Immigration and Social Issues, (5) Racism and Discrimination, (6) Personal Attacks. For the Portuguese variant, Galician (PT), we observe topics like (1) Feminism and Gender Advocacy, (2) Sexism and Gender Stereotypes, (3) Gender and Cars, (4) Misogyny and Gender Roles, (5) Racism and Discrimination, and (6) Patriarchy and Masculinity. Comparing both, the Galician (PT) topics were more narrowly focused on gender, while Galician (ES) maintained broader coverage. The lower variety of topics in the Galician (PT) data may be due to two main factors: the intrinsic characteristics of Galician as a lower-resource language (i.e. limitations in translation tools and natural language processing tools) and the effects of machine translation. These issues could lead to fewer and more repetitive topics, even though the dataset size is the same. Still, the frequent appearance of hate-related and gender-focused terms suggests that the main ideas from the original European Spanish data are preserved, especially around discrimination and online aggression.

The English dataset, developed independently, provided a comparison point. Its topics were: (1) Anti-Black Racism, (2) Hate and Extremism Rhetoric, (3) Misogyny and Homophobia, (4) White Supremacy and White Genocide Narrative, (5) Islamophobia and Anti-Muslim Sentiment, and (6) Slurs and Aggression. These topics include a mix of platform-specific behaviour (e.g., Wikipedia moderation), direct hate speech (e.g., racism, Islamophobia, misogyny), and broader social commentary (e.g., politics, pop culture). The English dataset shows that online hate can take different forms depending on cultural and social context. The inclusion of this sample helps us see how toxic language appears naturally in English, revealing both shared and language-specific patterns of online hostility.

Together, these topic clusters offer a way to explore hate speech across multiple languages. Since our dataset includes both translated and original texts, it supports cross-lingual comparisons and highlights how language, culture, and translation influence the way harmful speech is expressed and understood online.

\begin{figure}[ht]
    \centering
    \includegraphics[scale=0.55]{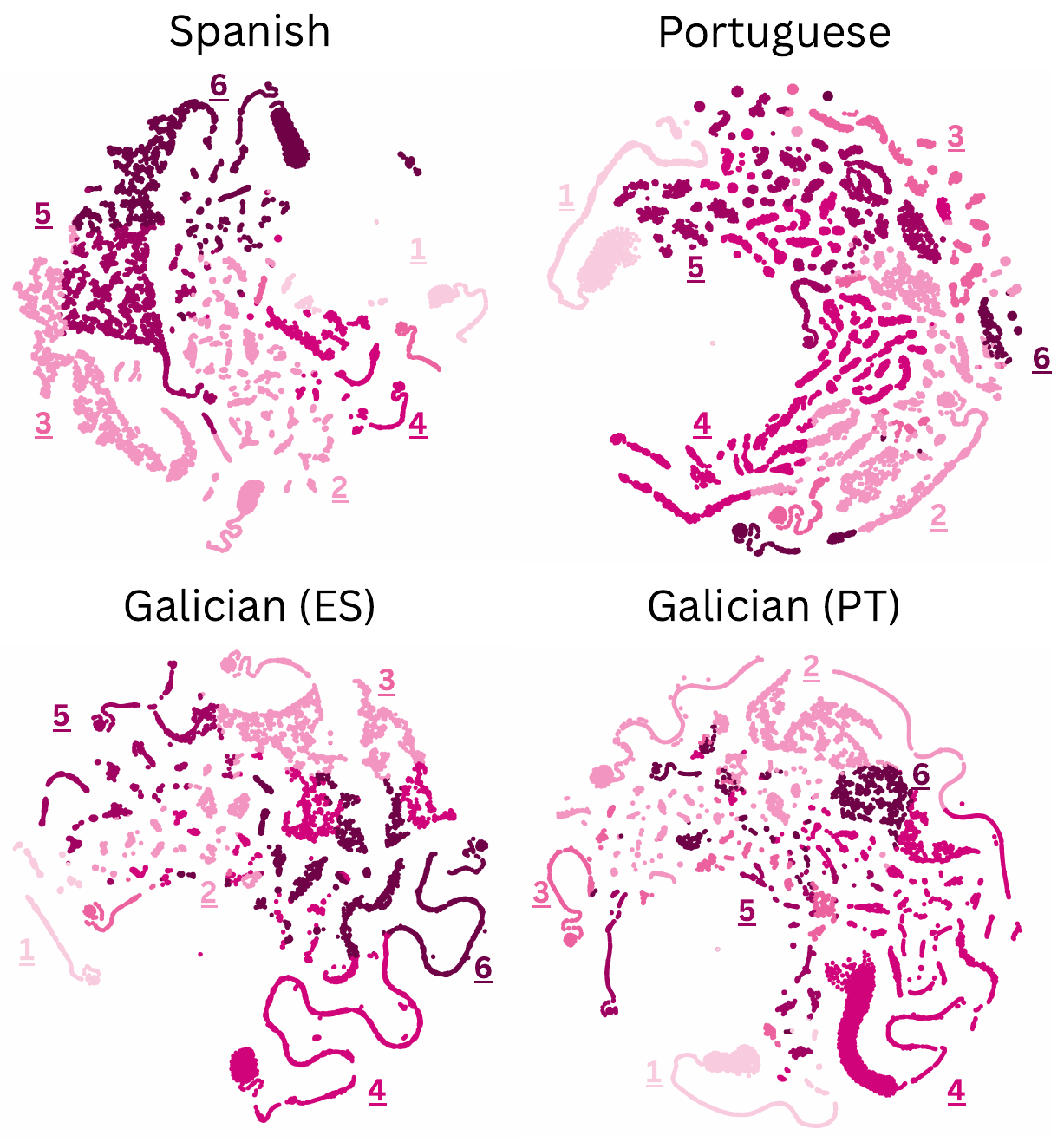}
    \caption{t-SNE visualisation of topics for Spanish, Portuguese, Galician (ES) and Galician (PT).}
    \label{fig:tsne}
\end{figure}

Next, we represent these topics using t-SNE. t-SNE is a dimensionality reduction algorithm, to visualise topic distributions. Figure~\ref{fig:tsne} shows how different types of hate speech cluster based on their semantic similarity. Each cluster represents a distinct topic, and the closer two clusters appear, the more similar their content. To improve the visualisation, we applied topic weights so that more prominent topics are more clearly represented. In the Spanish data, topics (1) and (4), related to feminism, gender, and emotional expression, appear close together on the right. Topics on racism and immigration, such as (2) and (3), are grouped near the bottom, reflecting their semantic overlap. On the left, clusters related to insults and slurs are also grouped together. Similar patterns are seen in the Portuguese and Galician datasets: gender-related topics are grouped, and topics on racism or immigration appear close together. While the clusters are distinguishable, their layout shows clear connections between related themes—such as immigration and racism, or gender and misogyny, rather than being randomly spread. This t-SNE visualisation helps reveal the underlying semantic structure of the hate speech topics

Based on the topic analysis carried out with BERTopic, there is a clear concentration of hate speech around topics related to sexism and misogyny. Although the dataset includes mentions of other issues such as immigration or racism, these appear with less frequency and intensity, suggesting a potential bias in the distribution of content toward gender-related matters. This finding highlights the need to expand the resources and corpora used for studying hate speech to more evenly represent other relevant dimensions, such as xenophobia, LGBTQ+phobia, aporophobia, or political hate \citep{curto-etal-2025-tackling}. Focusing solely on sexism as the main axis can lead to a partial view of the phenomenon, affecting both the automatic detection and the sociolinguistic understanding of online hate.

\subsection{Psycholinguistic analysis}

For our linguistic analysis, we conducted an analysis of emotions in our data, as well as the usage of pronouns and verb tenses.

\begin{figure}[ht]
    \centering
    \includegraphics[scale=0.37]{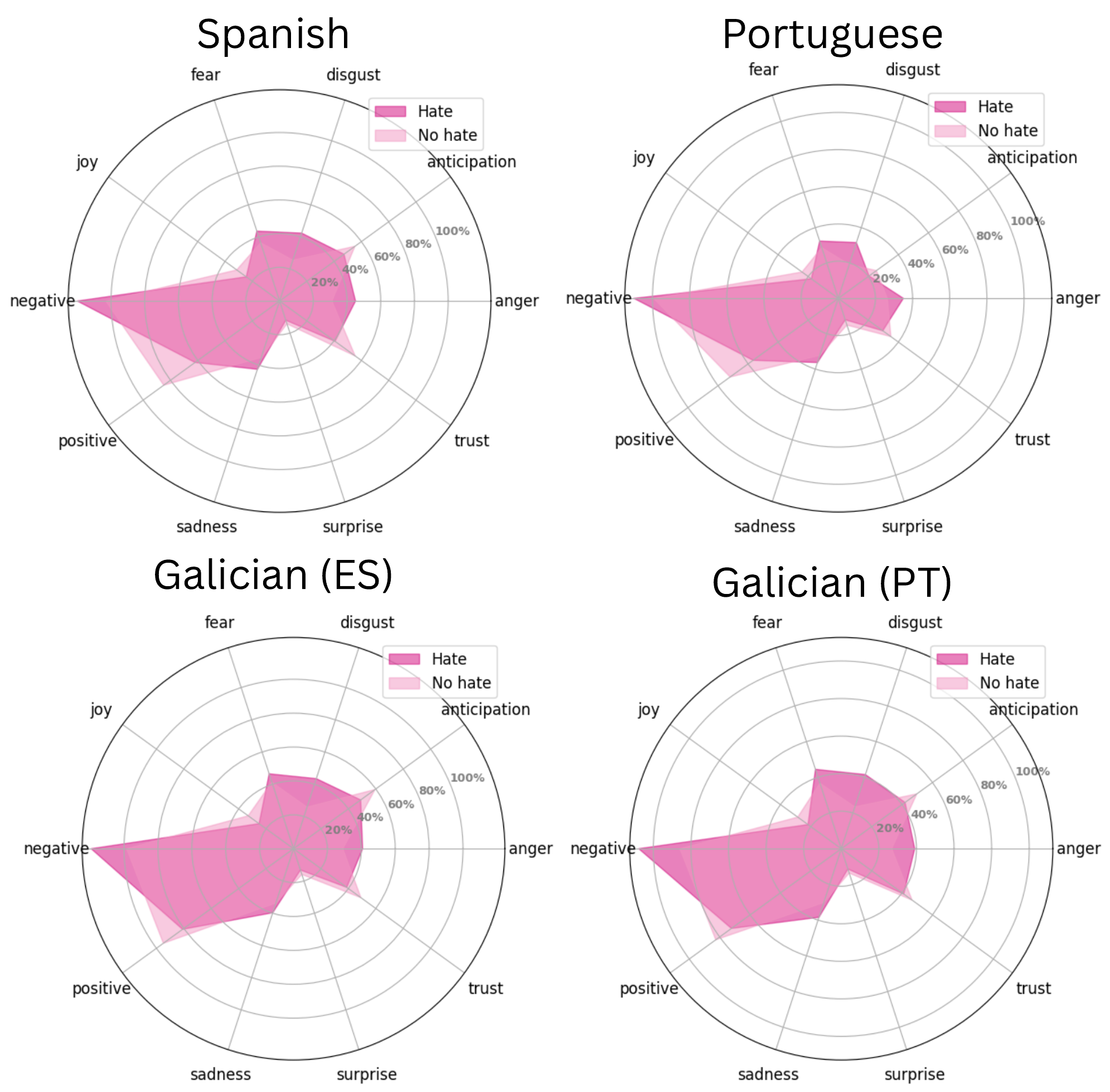}
    \caption{Radar plot showing the percentage of posts that contain a word associated with the Plutchik emotions for hate and non-hate data, for Spanish, Portuguese, Galician (ES) and Galician (PT).}
    \label{fig:pultchik}
\end{figure}

To analyse emotion, we  employed  the  Plutchik set  of  emotions \citep{plutchik1980general},  encompassing  eight  primary emotions (anger, fear, sadness, disgust, surprise, anticipation,  trust,  and  joy)  and  two  basic  sentiments  (positive and negative). To quantify emotion levels, we utilised the NRC emotion lexicon \citep{Mohammad13}, which comprises words linked to the Plutchik emotions. We  computed  the  percentage  of  posts  within each group (hate, no hate) that contained at least one word associated with the primary emotions and sentiments. We quantified the emotion levels for our four languages. As expected, the four datasets follow a similar pattern and distribution of emotions. Figure \ref{fig:pultchik}, shows that \textbf{negative} emotions are more frequent in hate speech posts, while \textbf{positive} emotions are more common in non-hateful content. Emotions such as \textbf{fear}, \textbf{disgust}, \textbf{anger}, and \textbf{sadness} appear more frequent  in  hate  messages. On the other hand,  emotions  like \textbf{trust} are  associated  with non-hate posts. Negative emotions are the most common in our dataset, likely due to the way hate speech data is collected—often using keywords and lexicons focused on negative terms. Emotions like \textbf{disgust} and \textbf{anger} show a strong contrast between hate and non-hate posts, appearing much more frequently in hate speech. The main difference across languages is seen in \textbf{anticipation} for Portuguese, where both hate and non-hate posts show notably lower levels compared to the other languages. This could be most likely because we are using European Portuguese, and NRC emotion lexicon in Portuguese was originally translated from English using Google Translate \citep{Mohammad13}, which might be biased towards Brazilian, limiting the lexicon for European Portuguese and limiting the discovering of this emotion. We had no control over this lexicon, as it is an existing resource, but based on our findings, future work should improve NRC emotion lexicon to reflect the languages variants. We highlight again, the importance to build resources that capture the variations among languages and dialects.

\begin{figure}
    \centering
    \includegraphics[scale=0.15]{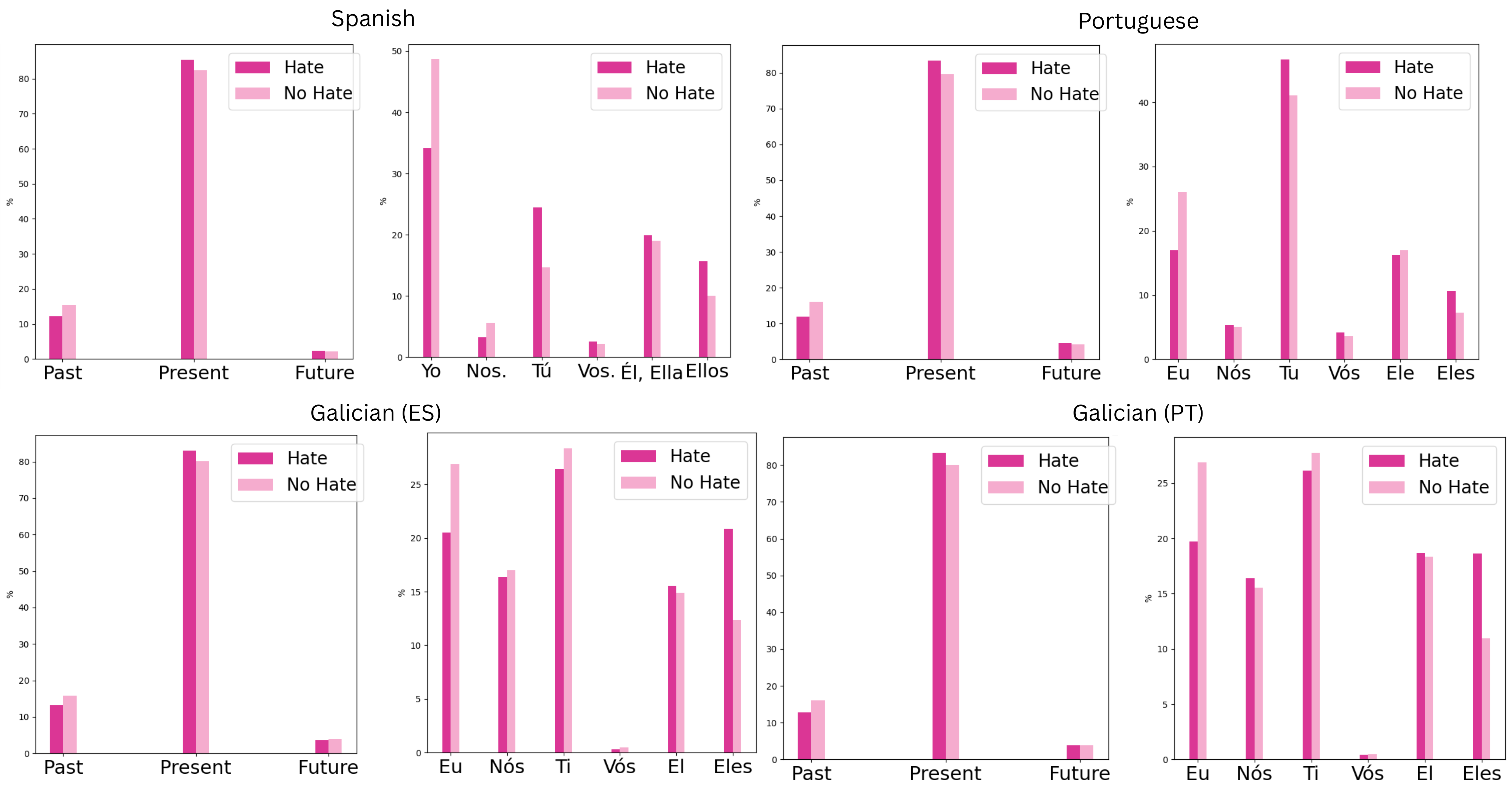}
    \caption{For each language, Spanish, Portuguese, Galician (ES) and Galician (PT), distribution of verb tenses in hate and non-hate posts (left). And distribution of pronouns in hate and non-hate speech posts (right).}
    \label{fig:verbs_pronouns}
\end{figure}

Regarding pronouns and verb tenses usage, we used spaCy with the Spanish (\texttt{es\_core\_news\_sm}) and Portuguese (\texttt{pt\_core\_news\_sm}) language models. Figure \ref{fig:verbs_pronouns} shows that, for all languages, there is a higher prevalence of present verb tenses in hate speech posts. Future tenses appeared rarely ($\sim$3\%), and past tenses less often ($\sim$20\%). This suggests that the higher prevalence of present tenses in hate speech posts are linked to direct attacks.

Regarding the usage of pronouns, for both Spanish and Portuguese datasets, non-hate speech post used more first-person singular pronouns, while hate speech posts used more second-person singular and third-person plural pronouns. Both ``you'' and ``they'' pronouns often indicate personal attacks and \textbf{othering} \citep{Rohleder2014}. Othering is a sociological concept where a group in labelled as fundamentally different, alien, or inferior to a dominant in-group, often involving pronouns like ``they''  to create a psychological and social divide. In Galician, we observed similar patterns, with a higher prevalence of first person singular pronouns in non-hate speech posts, and the third person plural pronoun being more frequent in hate speech posts. However, the second person singular pronoun showed less variation. Since no spaCy model exists for Galician, we used the Portuguese model, which may affect results.

These findings answer \textit{\textbf{RQ2}: Can synthetic data generation effectively compensate for the lack of annotated hate speech resources in European Portuguese and Galician?}. We show the effectiveness of synthetic data generation for low-resource languages like European Portuguese and Galician. Key hate speech themes and psycholinguistic patterns, such as negative emotions, present-tense usage, and othering pronouns, were preserved across translations. While Galician (PT) showed reduced topic variety, likely due to translation and tool limitations, the overall consistency suggests that synthetic datasets offer a practical solution for extending hate speech analysis to underrepresented Iberian languages.

\section{Experiments and results}\label{experiments}

In this section, we present our efforts to establish a common baseline for hate speech detection on social media for European Spanish, European Portuguese and Galician. To do so, we evaluate the proposed collections by benchmarking performance across several experimental settings using two large language models (\texttt{Llama 3.1} and \texttt{Nemo}) and a multilingual BERT model (\texttt{bert-base-multilingual-cased}). We conduct binary classification experiments under zero-shot, few-shot, and fine-tuning configurations in order to establish baseline results for future comparisons.

\subsection{Models}

We use two LLMs and one encoder-based model in our evaluation:

\begin{itemize}
    \item \textbf{Llama 3.1 8B} \citep{dubey2024llama3herdmodels}: Llama 3.1 8B is a pre-trained model, outperforming competing models such as Mistral 7B or Gemma 7B in tasks such as commonsense understanding, mathematical reasoning tasks and general tasks.

    \item \textbf{Nemo} \citep{mistral_nemo}: Nemo is an instruction-tuned model excelling in multilingual tasks and zero-shot scenarios, outperforming models like Mistral 7B in comprehension and domain-specific applications.
    
    \item \textbf{Multilingual BERT (\texttt{bert-base-multilingual-cased})} \citep{devlin-etal-2019-bert}: BERT multilingual base model is a encoder-only model, which has been pre-trained on 104 languages with Wikipedia data using masked language modelling. 
\end{itemize}

\subsection{Settings}

We perform experiments under the following configurations:

\begin{itemize}
    \item \textbf{Zero-shot}: The model is prompted with only the task description and input text, with no examples provided. Text generation is performed using the model's decoder with the following configuration: \texttt{max\_new\_tokens=256}, \texttt{temperature=0}, \texttt{top\_p=0.1}, and \texttt{top\_k=5}, ensuring low-variance, high-confidence completions.
    \item \textbf{Few-shot}: The model is given a small number of in-context examples (8 labelled instances) along with the task prompt. Text generation is performed using the same configuration as for zero-shot.
    \item \textbf{Fine-tuning}: The LLMs were fine-tuned on the training portion of each dataset and evaluated on held-out data. We fine-tuned for 3 epochs with a learning rate of $2 \times e^{-5}$, and batch size of 16. The AdamW optimiser and linear learning rate scheduler with warm-up were applied. Evaluation metrics include micro and macro-averaged F1-score. The BERT model was fine-tuned for 3 epochs with a learning rate of  $5 \times e^{-5}$, and batch size of 32. We set a weight decay of 0.18 and AdamW 8bit optimiser.
\end{itemize}

All experiments are conducted across five language variants: Spanish, Portuguese, Galician (ES) (Galician translated from Spanish), Galician (PT) (Galician translated from Portuguese), and English (an external dataset for cross-lingual validation). Every dataset was divided intro training and test sets, with a test size of 20\%. This multilingual evaluation allows us to assess model performance across closely related languages and translation directions, highlighting potential challenges in cross-lingual hate speech detection.

\subsection{Cross-lingual evaluation}

To further examine the generalisability of our models, we conduct cross-lingual evaluations using the fine-tuned models. Specifically, we evaluate how well a model fine-tuned on one language performs when applied directly to test sets in other languages, without additional adaptation. This setup enables zero-shot cross-lingual transfer learning through fine-tuned models.

For each language, we fine-tune BERT, Llama 3.1 and Nemo on its training split and test the resulting model on the other language test sets. This results in a cross-lingual evaluation matrix, where each cell corresponds to training in language $L_i$ and testing in language $L_j$. By evaluating performance across Spanish, Portuguese, Galician (ES), Galician (PT), and English, we assess cross-lingual transferability, with a focus on typologically related language pairs.

This analysis reveals the extent to which linguistic similarity and translation direction affect model performance in hate speech detection. It also provides insights into the robustness of our models and the utility of synthetic or translated data in low-resource settings.

\subsection{Results}

Table~\ref{tab:results} presents the performance of two multilingual models, Llama 3.1 and Nemo, across five language datasets (English, Spanish, Portuguese, and the two Galician variants) under three evaluation settings: zero-shot, few-shot, and fine-tuning. As expected, performance improves consistently across all languages when moving from zero-shot to fine-tuning, highlighting the importance of task-specific adaptation.

English yields the highest performance in all settings, which aligns with the availability of abundant training resources and benchmarks in this language. Spanish follows, benefiting from the large-scale dataset we constructed (+80k examples), enabling strong fine-tuned results (F1 micro $>$ 0.85). Portuguese and Galician show lower performance, particularly in the zero-shot and few-shot settings, which reflects the relative scarcity of data in LLMs in these variants. Notably, European Portuguese and Galician (both ES and PT variants) perform worse than Spanish, underscoring the challenges of low-resource settings.

The results emphasise the need for larger, more diverse multilingual hate speech datasets. While LLMs show encouraging transfer capabilities, fine-tuned models still significantly outperform zero-shot and few-shot settings, especially in underrepresented languages. Moreover, the variation between Galician (ES) and Galician (PT), despite sharing a common source, suggests that linguistic nuance and regional usage play a critical role in model generalisation. These findings reinforce the importance of both language-specific data curation and multilingual model training that captures dialectal diversity.

\begin{table}[ht]
    \centering    
    \renewcommand{\arraystretch}{1.1}
    \setlength{\tabcolsep}{7pt}
    \resizebox{\textwidth}{!}{
    \begin{tabular}{ll cccccc}
    \toprule
     & & \multicolumn{2}{c}{\textbf{{ Llama 3.1}}} & \multicolumn{2}{c}{\textbf{{Nemo}}}  & \multicolumn{2}{c}{\textbf{{BERT}}} \\
    \cmidrule(lr){3-4} \cmidrule(lr){5-6} \cmidrule(lr){7-8}
    \textbf{Language} & \textbf{Setting} & \textbf{{F1 Micro}} & \textbf{{F1 Macro}} & \textbf{{F1 Micro}} & \textbf{{F1 Macro}} & \textbf{{F1 Micro}} & \textbf{{F1 Macro}} \\
    \midrule
    English & \texttt{zero-shot}  & 0.7916 & 0.7643 & 0.7904 & 0.7647 & \multicolumn{2}{c}{--} \\
            & \texttt{few-shot} & 0.8044 & 0.7464 & 0.8092 & 0.7531 & \multicolumn{2}{c}{--} \\
            & \texttt{fine-tuned} & {0.8614} & {0.8240} & {0.8652} & {0.8334} & \textbf{0.9229} & \textbf{0.9048} \\
    \midrule
    Spanish & \texttt{zero-shot} & 0.7363 & 0.6854 & 0.7170 & 0.6848 & \multicolumn{2}{c}{--}  \\
            & \texttt{few-shot} & 0.7142 & 0.6780 & 0.7806 & 0.7236 & \multicolumn{2}{c}{--}  \\
            & \texttt{fine-tuned} & {0.8309} & {0.7575} & \textbf{0.8535} & \textbf{0.7981} & 0.8211 & 0.7583 \\
    \midrule
    Portuguese & \texttt{zero-shot} & 0.7339 & 0.6794 & 0.7380 & 0.6918 & \multicolumn{2}{c}{--}  \\
               & \texttt{few-shot} & 0.7178 & 0.6757 & 0.7697 & 0.7144 & \multicolumn{2}{c}{--}  \\
               & \texttt{fine-tuned} & {0.8113} & {0.7056} & \textbf{0.8262} & {0.7481} & 0.8174 & \textbf{0.7501} \\
    \midrule
    Galician (ES) & \texttt{zero-shot} & 0.7266 & 0.6501 & 0.7108 & 0.6644 & \multicolumn{2}{c}{--}  \\
                  & \texttt{few-shot} & 0.6868 & 0.6471 & 0.7469 & 0.6889 & \multicolumn{2}{c}{--}  \\
                  & \texttt{fine-tuned} & {0.8081} & {0.7168} & \textbf{0.8132} & \textbf{0.7351} & 0.8003 & 0.7274 \\
    \midrule
    Galician (PT) & \texttt{zero-shot} & 0.7311 & 0.6482 & 0.7222 & 0.6722 & \multicolumn{2}{c}{--}  \\
                  & \texttt{few-shot} & 0.7055 & {0.6602} & 0.7570 & 0.6954 & \multicolumn{2}{c}{--}  \\
                  & \texttt{fine-tuned} & {0.7803} & 0.5926 & \textbf{0.8131} & {0.7176} & 0.8047 & \textbf{0.7331} \\
    \bottomrule
    \end{tabular}
    }    
    \caption{Results.}
    \label{tab:results}
\end{table}

Table \ref{tab:crosslingual_results} presents the results of the cross-lingual evaluation, including our target languages and English. BERT performs well when tested on the same language it was trained on, consistently achieving the highest F1 scores in those settings (e.g., 0.82 micro-F1 for Spanish, 0.82 for Portuguese, and 0.80 for Galician-ES and Galician-PT). However, performance drops notably when transferring to other languages. For example, BERT trained on Spanish sees a decrease when applied to Galician (ES: 0.7819, PT: 0.7798), and a larger drop when tested on English (micro-F1: 0.7270). Spanish performs slightly better when transferring to Galician (ES) than to Galician (PT), suggesting closer alignment in syntax or vocabulary between Spanish and the Galician variant translated from Spanish. Similarly, Llama 3.1 and Nemo show strong performance in same-language settings for Spanish, achieving 0.83 (Llama) and 0.85 (Nemo) micro-F1 on Spanish test data. These models also generalise relatively well to related languages. For instance, Spanish-trained Llama achieves 0.7856 on Galician (ES) and 0.7816 on Galician (PT), while Nemo reaches 0.7968 and 0.7982 respectively. Although the performance gap between Galician (ES) and Galician (PT) is small, Galician (ES) tends to show slightly higher macro-F1 scores, particularly for Llama (0.6148 vs. 0.5995), indicating marginally better alignment with the Spanish-trained models. Overall, models trained on Spanish tend to transfer more effectively to Galician (ES) than Galician (PT), reflecting the influence of the translation direction in the Galician variants. This pattern is observed consistently across all three architectures.

Across all settings, BERT achieves its best performance when the training and testing languages match, regardless of the language, indicating its strong in-language specialisation. In contrast, Llama 3.1 and Nemo show a different pattern. When trained on Portuguese or either Galician variant, these models achieve their highest performance on the Spanish test set, rather than on their respective training languages. This suggests that Spanish serves as a strong target for cross-lingual generalisation in LLM-based models, possibly due to its broader data coverage or structural similarity to the other languages. For English, however, the pattern aligns with BERT: all models perform best when trained and tested on English. This reflects the well-resourced nature of English and the high alignment between its training and evaluation data.

\begin{table}[ht]
    \centering
    \renewcommand{\arraystretch}{1.1}
    \setlength{\tabcolsep}{7pt}
    \resizebox{\textwidth}{!}{
    \begin{tabular}{llcccccc}
    \toprule
    & & \multicolumn{2}{c}{\textbf{Llama 3.1}} & \multicolumn{2}{c}{\textbf{Nemo}} & \multicolumn{2}{c}{\textbf{BERT}} \\
    \cmidrule(lr){3-4} \cmidrule(lr){5-6} \cmidrule(lr){7-8}
    \textbf{Train} & \textbf{Test} & \textbf{F1 Micro} & \textbf{F1 Macro} & \textbf{F1 Micro} & \textbf{F1 Macro} & \textbf{F1 Micro} & \textbf{F1 Macro} \\
    \midrule

    \textbf{Spanish}        & \textbf{Spanish}        & \textbf{0.8309} & \textbf{0.7575} & \textbf{0.8535} & \textbf{0.7981} & \textbf{0.8211} & \textbf{0.7583}  \\
                   & Portuguese     & 0.7956 & 0.6434 & 0.8197 & 0.7170 & 0.7906 & 0.6860 \\
                   & Galician (ES)  & 0.7856 & 0.6148 & 0.7968 & 0.6713 & 0.7819 & 0.6688 \\
                   & Galician (PT)  & 0.7816 & 0.5995 & 0.7982 & 0.6716 & 0.7798 & 0.6608 \\
                   & English        & 0.7929 & 0.6792 & 0.7996 & 0.7103 & 0.7270 & 0.4962  \\
    \midrule

    \textbf{Portuguese}     & Spanish        & \textbf{0.8133} & 0.7120 & \textbf{0.8335} & \textbf{0.7598} & 0.8078 & 0.7278 \\
                   & \textbf{Portuguese}     & 0.8113 & 0.7056 & 0.8262 & 0.7481 & \textbf{0.8174} & \textbf{0.7501} \\
                   & Galician (ES)  & 0.7851 & 0.6266 & 0.7978 & 0.6717 & 0.7829 & 0.6922 \\
                   & Galician (PT)  & 0.7851 & 0.6265 & 0.7979 & 0.6737 & 0.7892 & 0.7061 \\
                   & English        & 0.8032 & \textbf{0.7236} & 0.8000 & 0.7148 & 0.7546 & 0.6274  \\
    \midrule

    \textbf{Galician (ES)}  & Spanish        & \textbf{0.8181} & \textbf{0.7514} & \textbf{0.8297} & \textbf{0.7707} & 0.7874 & 0.7163 \\
                   & Portuguese     & 0.8098 & 0.7228 & 0.8195 & 0.7408 & 0.7884 & 0.7072 \\
                   & \textbf{Galician (ES)}  & 0.8081 & 0.7168  & 0.8132 & 0.7351 & \textbf{0.8003} & \textbf{0.7274} \\
                   & Galician (PT)  & 0.8033 & 0.7035 & 0.8097 & 0.7247 & 0.7998 & 0.7203 \\
                   & English        & 0.8010 & 0.7115 & 0.8076 & 0.7442 & 0.7395 & 0.5791  \\
    \midrule

    \textbf{Galician (PT)}  & Spanish        & \textbf{0.7893} & \textbf{0.6299} & \textbf{0.8265} & \textbf{0.7444} & 0.7969 & 0.7175 \\
                   & Portuguese     & 0.7795 & 0.5883 & 0.8154 & 0.7202 & 0.7964 & 0.7176 \\
                   & Galician (ES)  & 0.7828 & 0.6024 & 0.8136 & 0.7177 & 0.7984 & 0.7237 \\
                   & \textbf{Galician (PT)}  & 0.7803 & 0.5926 & 0.8131 & 0.7176 & \textbf{0.8047} & \textbf{0.7331} \\
                   & English        & 0.7678 & 0.6248 & 0.8095 & 0.7425 & 0.7395 & 0.5674  \\
    \midrule

    \textbf{English}        & Spanish        & 0.7655 & 0.6003 & 0.7657 & 0.6702 & 0.7353 & 0.5762  \\
                   & Portuguese     & 0.7512 & 0.6167 & 0.7606 & 0.6563 & 0.7316 & 0.5684 \\
                   & Galician (ES)  & 0.7513 & 0.5669 & 0.7456 & 0.6232 & 0.7281 & 0.5660 \\
                   & Galician (PT)  & 0.7525 & 0.5529 & 0.7600 & 0.6304 & 0.7353 & 0.5613  \\
                   & \textbf{English}        & \textbf{0.8614} & \textbf{0.8240} & \textbf{0.8652} & \textbf{0.8334} & \textbf{0.9229} & \textbf{0.9048}  \\
    \bottomrule
    \end{tabular}
    }
    \caption{Cross-lingual evaluation results (micro and macro F1 scores) for models fine-tuned on each training language and tested on multiple target languages.}
    \label{tab:crosslingual_results}
\end{table}

\begin{figure}
    \centering
    \includegraphics[scale=0.45]{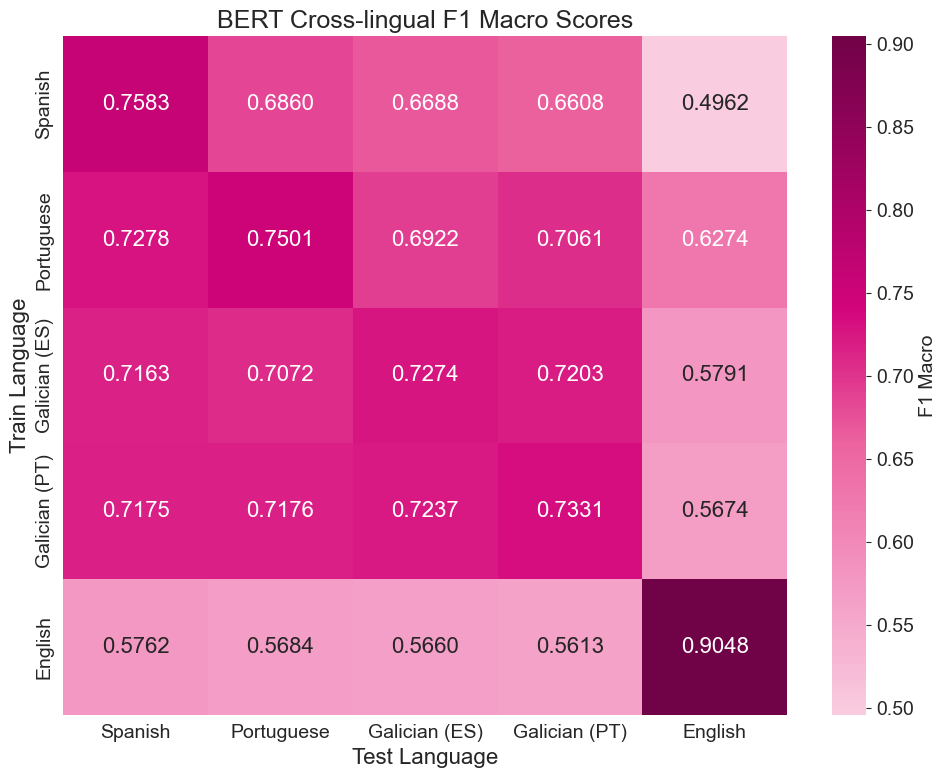}
    \caption{Cross-lingual evaluation of BERT-based classifiers fine-tuned on each training language (rows) and evaluated on each target language (columns). Values represent F1 scores (macro-averaged) for the binary classification task of hate speech detection.}
    \label{fig:cross-lingual-matrix}
\end{figure}

Figure \ref{fig:cross-lingual-matrix} further illustrates the cross-lingual performance of the BERT-based classifiers fine-tuned on a source language and evaluated across multiple target languages. The highest F1 Macro scores are observed along the diagonal, indicating that the model performs best when evaluated on the same language it was trained on. Notably, the model trained on English exhibits the strongest self-performance (F1 = 0.9048), followed by Spanish (F1 = 0.7583). Cross-lingual generalisation is generally strong among the closely related Iberian languages, suggesting that linguistic similarity contributes to effective transfer. For instance, results show that Spanish-trained models perform best on Galician (ES variant), achieving an F1 score of 0.6688, while Portuguese-trained models perform best on Galician (PT variant) with an F1 of 0.7061. This suggests that Galician varieties benefit most from training on their closest high-resource counterparts, Spanish for Galician (ES), and Portuguese for Galician (PT), highlighting the importance of linguistic proximity. 

Interestingly, models trained on Galician (both variants) tend to generalise better to Spanish and Portuguese than the reverse. For example, Galician (PT) model achieve F1 scores of 0.7175 on Spanish and 0.7176 on Portuguese, which are competitive or superior compared to the inverse direction (0.6608 and 7061, respectively). This suggests that training on Galician, despite being a lower-resource language, may capture generalisable patterns that benefit related Iberian languages.

Cross-lingual performance between other Iberian languages also remains relatively high, while a more significant drop is observed when transferring between Iberian and English, indicating reduced effectiveness across more distant language families. Notably, the Iberian language datasets (Spanish, Portuguese, and both Galician variants) were derived from a shared source and translated across languages, whereas the English dataset is independent, which may also explain the relatively lower cross-lingual transfer performance between English and the Iberian languages. These results underscore the importance of language similarity and data alignment in multilingual model generalisation.

These results address \textit{\textbf{RQ3}: How well do state-of-the-art large language models perform in detecting hate speech across these low-resource linguistic varieties?} Our results highlight that state-of-the-art LLMs like Llama 3.1 and Nemo demonstrate promising performance in hate speech detection across low-resource varieties, particularly when fine-tuned. Performance improves consistently from zero-shot to few-shot to fine-tuned settings, highlighting the importance of low-resourced data. While English and Spanish benefit from stronger baselines due to LLMs having seen more data in these languages, Portuguese and Galician variants show competitive results when fine-tuned, despite their low-resource status. However, performance in zero and few-shot configurations remains notably lower, underscoring the continued challenges LLMs face in generalising to underrepresented language varieties without targeted training data.

Moreover, our results answer \textit{\textbf{RQ4}: Does accounting for Galician's internal variation (Spanish-like vs. Portuguese-like) affect model performance?} The internal variation within Galician clearly impacts model performance. Although both Galician (ES) and Galician (PT) come from the same original content, they produce different results, especially in macro F1 scores. This suggests that regional differences in vocabulary or grammar influence how well models generalise. These findings highlight the importance of considering dialectal differences when building or testing models for low-resource languages, since even small shifts between Spanish-like and Portuguese-like features can impact classification accuracy. Cross-lingual evaluation shows that, generally, models trained on Spanish perform better on Galician (ES), while models trained on Portuguese transfer better to Galician (PT), confirming the linguistic affinity within each variant. These differences emphasise the importance of considering internal dialectal variation when developing or evaluating models for low-resource languages, as even subtle shifts in linguistic alignment can affect classification accuracy and model robustness.

\section{Conclusion}\label{conclusion}

In this work, we addressed the scarcity of hate speech resources for low-resource Iberian languages and their linguistic varieties by compiling a meta-collection of hate speech datasets for European Spanish. We reformatted these datasets into a standardised structure with harmonised labels and metadata, facilitating interoperability and cross-lingual analysis. This meta-collection served as the foundation for translating and generating aligned datasets in European Portuguese and both Spanish and Portuguese varieties of Galician.

We performed a comprehensive analysis of these datasets through unsupervised topic modelling, emotion and tense profiling, and visualisation techniques (t-SNE clustering and word clouds), uncovering cross-lingual patterns and genre-specific tendencies in hate speech. This analysis revealed important linguistic and sociocultural dimensions that are often flattened in monolingual or English-centric studies.

Our benchmarking of state-of-the-art large language models, including Llama 3.1 and Nemo, across zero-shot, few-shot, and fine-tuned scenarios shows that while LLMs exhibit promising generalisation capabilities. But fine-tuning with high-quality, language-specific data remains essential for achieving robust performance, particularly in low-resource contexts. These findings underscore the value of investing in curated, balanced datasets and underscore the limitations of relying solely on pretrained multilingual models without adaptation. 

Regarding the cross-lingual results, our findings highlight the importance of training directly in the target language, as this consistently yields the best performance. However, when this is not possible, using a closely related language proves beneficial as models trained on linguistically similar languages tend to perform better than those trained on more distant ones. This reinforces the role of language proximity in cross-lingual transfer and underscores the need for language-specific resources when tackling hate speech detection.

Overall, our work highlights the need for multilingual, variety-aware approaches to hate speech detection. As LLMs continue to play a central role in content moderation and linguistic analysis, ensuring their fairness and linguistic inclusivity is critical. Future research should extend this framework to other underrepresented languages and varieties, mitigate potential translation biases, and expand the scope of hate speech analysis to include xenophobia, LGTBphobia, political hate, and other sociolinguistic dimensions. By doing so, we can move toward a more comprehensive and equitable understanding of harmful discourse online.

\section*{Declarations}

\begin{itemize}
\item Funding: Open Access funding provided thanks to the CRUE-CSIC agreement with Springer Nature. The authors thank the funding from the Horizon Europe research and innovation programme under the Marie Skłodowska-Curie Grant Agreement No. 101073351.  The first and third authors also thank the financial support supplied by the grant PID2022-137061OB-C21 funded by MI-CIU/AEI/10.13039/501100011033 and by “ERDF/EU”. The authors also thank the funding supplied by the Consellería de Cultura, Educación, Formación Profesional e Universidades (accreditations ED431G 2023/01 and ED431C 2025/49) and the European Regional Development Fund, which acknowledges the CITIC, as a center accredited for excellence within the Galician University System and a member of the CIGUS Network, receives subsidies from the Department of Education, Science, Universities, and Vocational Training of the Xunta de Galicia. Additionally, it is co-financed by the EU through the FEDER Galicia 2021-27 operational program (Ref. ED431G 2023/01).
\item Conflict of interest:  We declare that we have no conflict of interest as defined by Springer or other interests
that might be perceived to influence the results and/or discussion reported in this paper.
\item Data availability: Data is available upon request, and we have adhered to strict ethical standards, particularly those about the ethical development of AI (\url{https://huggingface.co/datasets/irlab-udc/MetaHateES}).
\item Materials availability: Materials are available in the data and code availability items.
\item Code availability: The code to replicate the meta-collection construction, translation, analysis and running the baselines is available here \url{https://github.com/palomapiot/mlhate}.
\item Author contribution: The first author contributed to the following aspects: conceptualisation, implementation, investigation, writing, and supervision. The second and third authors contributed to conceptualisation and supervision.
\end{itemize}


\bibliography{sn-bibliography}

\end{document}